\pdfoutput=1

\documentclass[11pt]{article}

\usepackage[]{acl}

\usepackage{times}
\usepackage{latexsym}
\usepackage[T1]{fontenc}\usepackage[utf8]{inputenc}
\usepackage{microtype}

\usepackage{amsmath,amsfonts,bm}

\def\eqref#1{equation~\ref{#1}}
\def\1{\bm{1}}

\DeclareMathAlphabet{\mathsfit}{\encodingdefault}{\sfdefault}{m}{sl}
\SetMathAlphabet{\mathsfit}{bold}{\encodingdefault}{\sfdefault}{bx}{n}

\usepackage{hyperref}
\usepackage{url}
\usepackage{booktabs}       % professional-quality tables
\usepackage{amsfonts}       % blackboard math symbols
\usepackage{nicefrac}       % compact symbols for 1/2, etc.
\usepackage{xcolor}         % colors
\usepackage{enumitem}
\newcommand{\xhdr}[1]{{\noindent\bfseries #1}.}
\usepackage{subfigure}
\usepackage{graphicx}
\usepackage{multirow}
\usepackage{utfsym}
\usepackage{diagbox}
\usepackage{colortbl}
\usepackage[most]{tcolorbox}
\usepackage{caption}
\usepackage{wrapfig}
\definecolor{mypink}{rgb}{.99,.91,.95}
\definecolor{mygreen}{rgb}{.9,.99,.9}
\definecolor{mygray}{gray}{.9}
\usepackage{marvosym}
\usepackage{framed}
\usepackage{mdframed}
\usepackage{pifont}
\usepackage{CJKutf8}
\usepackage{fontawesome}

\newcommand{\hide}[1]{}

\title{LongAlign: A Recipe for Long Context Alignment\\
Encompassing Data, Training, and Evaluation}

\title{LongAlign: A Recipe for Long Context Alignment\\
of Large Language Models}

\author{%
  Yushi Bai$^{\ddagger\dagger}$, Xin Lv$^{\S}$, Jiajie Zhang$^{\ddagger\dagger}$, Yuze He$^\ddagger$, Ji Qi$^{\ddagger\dagger}$, \\ 
  \textbf{Lei Hou$^\ddagger$, Jie Tang$^\ddagger$, Yuxiao Dong$^\ddagger$, Juanzi Li$^\ddagger$} \\
  $^\ddagger$Tsinghua University
  \quad
  $^\S$Zhipu.AI
}

\begin{document}
\maketitle

\renewcommand{\thefootnote}{\fnsymbol{footnote}}
    \footnotetext[2]{Work done when YB, JZ, and JQ interned at Zhipu.AI. %Email: \texttt{bys22@mails.tsinghua.edu.cn}
    }
\renewcommand{\thefootnote}{\arabic{footnote}}

\thispagestyle{plain}
\pagestyle{plain}

\begin{abstract}

Extending large language models to effectively handle long contexts requires instruction fine-tuning on input sequences of similar length. 
To address this, we present LongAlign---a recipe of the instruction data, training, and evaluation for long context alignment. 
First, we construct a long instruction-following dataset using Self-Instruct. 
To ensure the data diversity, it covers a broad range of tasks from various long context sources. 
Second, we adopt the packing and sorted batching strategies to speed up supervised fine-tuning on data with varied length distributions. 
Additionally, we develop a loss weighting method to balance the contribution to the loss across different sequences during packing training. 
Third, we introduce the LongBench-Chat  benchmark for evaluating instruction-following capabilities on queries of 10k-100k in length.
Experiments show that LongAlign outperforms existing recipes for LLMs in long context tasks by up to 30\%, while also maintaining their proficiency in handling short, generic tasks.
The code, data, and long-aligned models are open-sourced at \url{https://github.com/THUDM/LongAlign}.

\end{abstract}

\section{Introduction}
\label{sec:intro}

\begin{figure}[t]
    \centering
    \includegraphics[width=\linewidth]{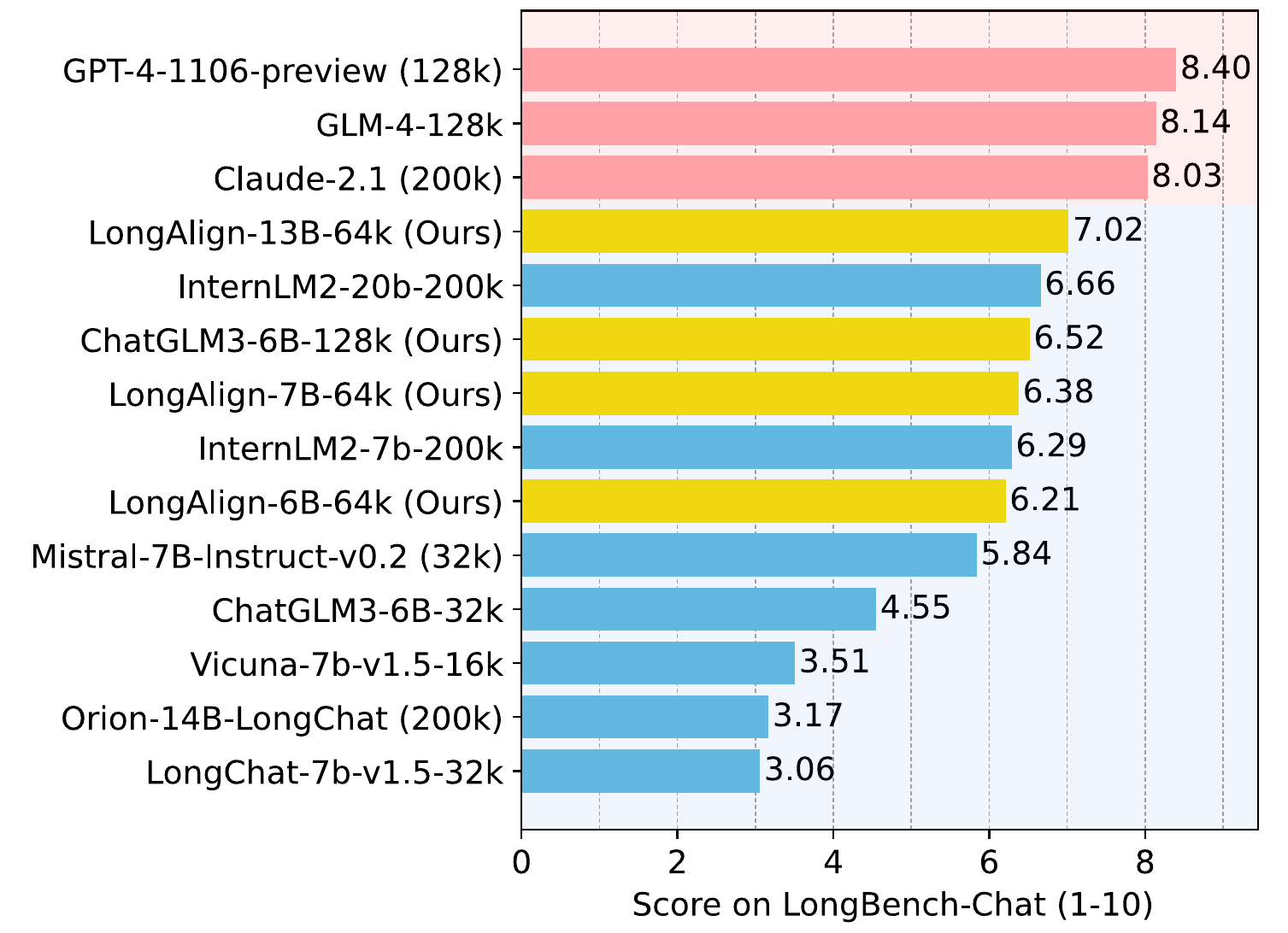}
    \caption{Test results on LongBench-Chat, which contains real-world queries of 10k-100k in length\protect\footnotemark.}
    \label{fig:leaderboard}
\end{figure}

Large language models (LLMs) with large context windows facilitate tasks such as summarization, question answering on long text and code~\cite{bai2023longbench}. 
Importantly, they may form the foundational support for life-long conversations and complex agent scenarios~\cite{xiao2023efficient,liu2023agentbench}. 
Existing works to build long-context LLMs predominantly focus on context extension~\cite{chen2023extending,xiong2023effective,peng2023yarn}, that is, position encoding extension and continual training on long text. 

In this work, we instead focus on the perspective of long context alignment, i.e., instruction fine-tuning LLMs to handle long user prompts. 
However, several challenges are required to address. 
First, there is an absence of long instruction-following datasets for supervised fine-tuning (SFT), and by extension the lack of methods for constructing such data. 
Second, the varied length distribution of long-context data drastically reduces the training efficiency of traditional batching methods in a multi-GPU setup, as GPUs processing shorter inputs have to stay idle until those handling longer inputs complete their tasks. 
Third, there is a crucial need for a robust benchmark to evaluate LLMs' long-context capacities against real-world queries.

\footnotetext{LongAlign-6B-64k, LongAlign-7B-64k and LongAlign-13B-64k are trained based on ChatGLM3-6B, Llama-2-7B and Llama-2-13B, respectively.}

To address them, we present the \textbf{LongAlign} recipe, covering data, efficient training, and evaluation, respectively. 
\textit{Data-wise}, to construct a diverse long instruction-following dataset, we collect long sequences from nine sources and use Self-Instruct~\cite{selfinstruct} to generate 10k instruction data of 8k-64k length. 

\textit{Training-wise}, to address the inefficiency under uneven batching, we adopt the packing strategy~\cite{krell2021efficient} that packs sequences together up to the maximum length before dispatching them to GPUs. 
However, we identified a bias in loss averaging during this packing training, as packs containing different numbers of sequences are assigned equal weight in the final loss calculation. 
To mitigate this bias, we propose a loss weighting strategy to balance contributions to the loss across different sequences.
In addition, we introduce a sorted batching method that groups sequences of similar lengths to reduce the intra-batch idle time.

\textit{Evaluation-wise}, we develop LongBench-Chat, a benchmark compromising open-ended questions of 10k-100k length annotated by Ph.D. students. 
It covers diverse aspects of instruction-following abilities such as reasoning, coding, summarization, and multilingual translation over long contexts. 
GPT-4~\cite{GPT-4} is employed to score the machine-generated responses based on our annotated groundtruths and few-shot scoring examples.

Extensive experiments show that LongAlign effectively aligns models to handle contexts of up to 64k tokens in length while maintaining their performance on general tasks without degradation. 
In addition, we have the following findings: 
\begin{itemize}[itemsep=0pt, leftmargin=*]
    \item \textbf{Impact of Data Quantity and Diversity}: 
    Both the quantity and the diversity of the long instruction data significantly influence the aligned model's ability to handle long contexts, impacting final performance by up to 30\%. 
    
    \item \textbf{Benefits of Long Instruction Data}: 
    The amount of long instruction data positively affects the performance on long-context tasks while does not hurt the models' short-context handling capacities. 
    
    \item \textbf{Effectiveness of Training Strategies}: 
    The packing and sorted batching strategies adopted can accelerate training by over 100\% without  performance compromise. 
    Furthermore, the proposed loss weighting technique improves long context performance by 10\%.
\end{itemize}

\section{Related Work}
\label{sec:related}

\xhdr{Long Context Scaling}
Long context scaling aims to expand the limited context length of existing LLMs to support long context tasks~\cite{xiong2023effective}. The current methods for long context scaling can be divided into two categories: those that require fine-tuning or continual training on longer sequences and those that do not.
Methods that do not require fine-tuning often employ techniques such as sliding window attention~\cite{han2023lm,xiao2023efficient} or neighboring token compression~\cite{jiang2023longllmlingua,zhang2024soaring,jin2024llm} to handle the positional O.O.D. problem in attention computation for long contexts.
These methods, although capable of extending the context length of LLMs in a plug-and-play manner, still cannot match the performance of the fine-tuned approaches.
Prominent fine-tuned approaches for long context scaling~\cite{chen2023extending,peng2023yarn,xiong2023effective,chen2023longlora,zhu2023pose,fu2023longdata} typically involve position encoding extension and continual pretraining on longer sequences.

\xhdr{LLM Alignment}
Following the previous steps of long context scaling, it is vital to also align the model with instruction-following data to ensure that it can interact with various user requests in a chat interface~\cite{wang2023how}.
This phase, often referred to as supervised fine-tuning or instruction-tuning, has been extensively studied in short context scenarios~\cite{selfinstruct,alpaca,wang2023how,tunstall2023zephyr}.
However, the introduction of long sequences presents unique challenges in terms of data, training methods, and evaluation for alignment.
\citet{xiong2023effective} proposes generating long instruction data by concatenating short instruction data, yet their dataset and model weight are not open-sourced.
On the other hand, while \citet{chen2023longlora} has made their long instruction data, LongAlpaca-12k, available and employed LoRA~\cite{hu2022lora} for efficient fine-tuning, it lacks in-depth discussion and comparative analysis of the influence of data and training methodologies.
Our work aims to find an optimal solution for supervised (full parameter) fine-tuning on long context with full attention, by tuning data, training methods, and evaluating the aligned models on a wide range of tasks.

\section{LongAlign}
\label{sec:method}
In this section, we discuss the methodology in LongAlign, involving the data construction process, training method, and evaluation benchmark.

\begin{figure}[t]
    \centering
    \includegraphics[width=\linewidth]{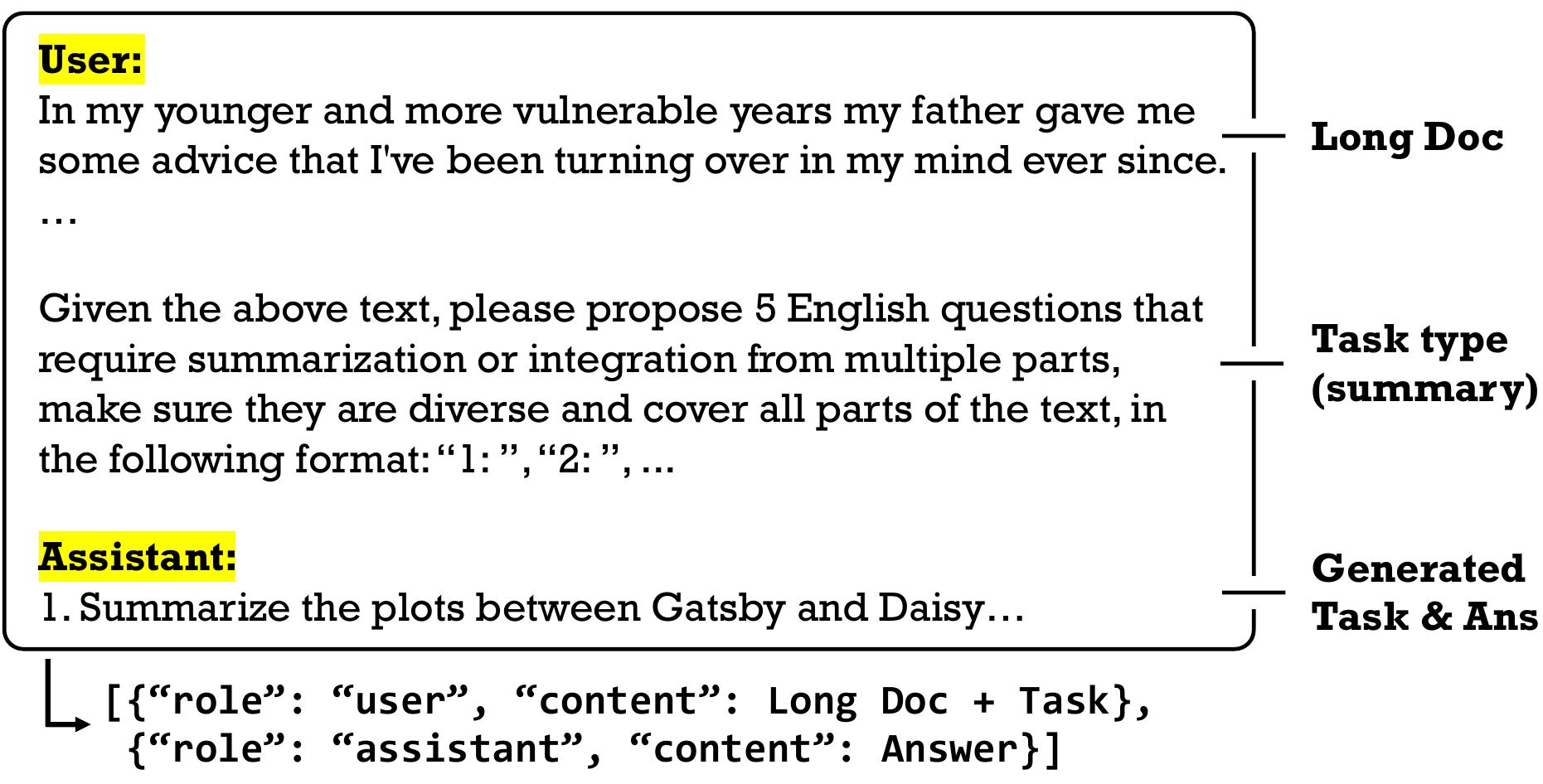}
    \caption{Data construction example.}
    \label{fig:inst}
\end{figure}

\begin{figure*}
    \centering
    \includegraphics[width=\linewidth]{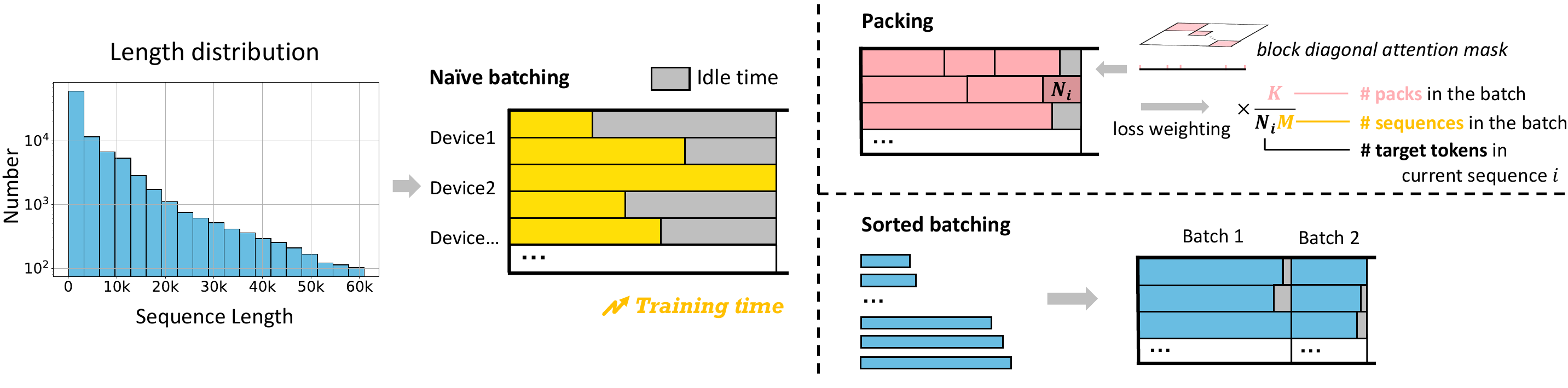}
    \caption{Under a long-tailed data length distribution, packing or sorted batching can reduce idle time and speed up the training process. Loss weighting is required during packing to balance the loss contribution across sequences.}
    \label{fig:method}
\end{figure*}

\subsection{Preliminary}
Large language models can learn alignment by supervised fine-tuning on high-quality pairs of instruction $x$ and response $y$~\cite{ouyang2022training,chung2022scaling}. During training, the instruction and response are typically concatenated to form a sequence $[x,y]$, which is then processed through an auto-regressive language model $\pi$ to maximize the probability $P_\pi(y|x)$.
The loss is similar to a language modeling loss, while only accounting for the loss associated with the tokens in $y$ (target tokens):
\begin{equation}
    \mathcal{L}([x, y]) = -\sum_{i=1}^{|y|}\log P_\pi(y_i\,|\,[x, y_{<i}]).
\end{equation}

\subsection{Dataset Construction}
\label{sec:data}
Long instruction data typically involves a long context material, such as a book, an extensive document, or a lengthy code, accompanied by a task query that requires summarizing, reasoning, or computing based on the material.
During construction, we first collect long articles and documents from 9 varied sources, covering books, encyclopedias, academic papers, codes, etc.
We then employ Claude 2.1~\cite{claude} to generate tasks and answers according to a given long context, as illustrated in Figure~\ref{fig:inst}.
To foster a diverse range of generated tasks, we incorporate task type descriptions into the prompts, such as queries for summaries, information extraction, reasoning, etc. 
Using this methodology, we create tasks and answers for 10k lengthy texts, yielding a total of 10k instances of supervised data, of which 10\% is in Chinese.
The length of these data ranges from 8k to 64k, measured by ChatGLM tokenizer~\cite{zeng2022glm} due to its higher compression rate for Chinese characters.
Details regarding the prompts and the data construction process can be found in Appendix~\ref{app:dataset}.

\subsection{Efficient Long-Context Training}
To ensure that the model retains the ability to handle both long and short texts (general capability) after SFT, we mix the long instruction data with a general instruction dataset for training.
The mixture of a large amount of general short data with a relatively smaller amount of long instruction data results in a long-tail data length distribution. As shown in Figure~\ref{fig:method} left, the majority of the data falls within the 0-8k length range, while the remaining data is fairly evenly distributed in the 8k-64k length interval.
Under this distribution, during training, a data batch typically contains mostly short data, yet these batches also include a few longer texts which necessitate much more computation times, resulting in considerable idle times.
To minimize these idle times, the most effective approach is to concatenate or sort the data in a manner that ensures a more uniform length and computational time within each batch.
Bearing this in mind, we explore two training methods, namely packing and sorted batching.

\xhdr{Packing} It involves concatenating data of varying lengths together until reaching the maximum length.
The resulting packed data, whose lengths are generally close to the maximum length, are then batched and processed on multi-GPUs.
This approach effectively minimizes the idle time within each batch, as depicted in the upper right of Figure~\ref{fig:method}.
Additionally, to prevent cross-contamination between different sequences within the same pack during self-attention calculation, we pass a list containing the starting and ending positions of different sequences and utilize the \texttt{flash\_attn\_varlen\_func} from FlashAttention 2~\cite{dao2022flashattention,dao2023flashattention2}, which supports efficient computation of block diagonal attention (see Appendix~\ref{app:training} for more details).
It requires less computation and IO time compared to the traditional use of a 2D attention mask.

However, we notice that the packing strategy leads to a bias towards longer sequences and sequences containing more target tokens.
This is because different packs, each contributing equally to the final loss, contain varying numbers of sequences with different numbers of target tokens. Consequently, when calculating the mean loss for each batch, sequences in packs with fewer sequences (typically the longer ones) or those containing more target tokens, have a greater influence on the final loss.
Formally, consider $M$ sequences packed into a batch of $K$ packs where the $i$-th pack consists of the sequences with indices in $[P_{i-1}, P_i)$, thus it holds that $P_0=1, P_K=M+1$. Let $L_i$ denote the total summation of loss over $N_i$ target tokens in the $i$-th sequence. If we weigh each sequence equally, the loss should be
\begin{equation}
    \mathcal{L} = \frac{1}{M}\sum_{i=1}^{M}\frac{L_i}{N_i},
\label{eq:loss}
\end{equation}
while the loss calculated under packing is 
\begin{equation}
    \mathcal{L}' = \frac{1}{K}\sum_{k=1}^{K}(\sum_{i=P_{k-1}}^{P_{k}-1}L_i / \sum_{i=P_{k-1}}^{P_{k}-1}N_i) \neq \mathcal{L}.
\label{eq:loss'}
\end{equation}
Compared with Eq.~\ref{eq:loss}, this equates to assigning a weight of $(N_j/\sum_{i=P_{k-1}}^{P_{k}-1}N_i)$ to sequence $j$ in the loss, i.e., in favor of sequences with more target tokens and sequences in smaller packs.
To address this inequality, we propose to scale the loss in the $i$-th sequence by $K/(N_iM)$ and instead take the sum of the scaled loss on each pack, which results in an equal loss to Eq.~\ref{eq:loss}:
\begin{equation}
    \mathcal{L}' = \frac{1}{K}\sum_{k=1}^{K}(\sum_{i=P_{k-1}}^{P_{k}-1}\frac{L_iK}{N_iM})
    = \frac{1}{K}\sum_{i=1}^{M}\frac{L_iK}{N_iM}=\mathcal{L}.
\end{equation}
As demonstrated in our experimental section, the loss weighting strategy results in a 10\% improvement in downstream tasks.

\xhdr{Sorted batching}
We also consider an efficient sorted batching strategy for training (lower right of Figure~\ref{fig:method}). To ensure that the sequences within each batch are of similar lengths, we sort the data by length and select a random consecutive group of data for each batch, with no repetition.
However, this strategy inevitably introduces a bias in the data distribution across different batches, where batches consist either of all long sequences or all short sequences. 
This can be potentially disastrous for SGD optimization.
In our experiments, we observe that sorted batching significantly accelerates the process without a noticeable negative impact on performance. This might be attributed to our use of large gradient accumulation steps and the strong adaptability of the optimizer.

\subsection{LongBench-Chat}
Although there are existing benchmarks for evaluating LLMs' long context understanding~\cite{an2023leval,bai2023longbench,li2023loogle}, they do not focus on assessing their instruction-following capability under long context. Furthermore, their reliance on automatic metrics for evaluation limits the assessment of aligned models' longer and more diverse outputs to real-world queries, and how their responses align with human preference.

To this end, we propose LongBench-Chat, which includes 50 long context real-world queries ranging from 10k to 100k in length, covering various key user-intensive scenarios such as document QA, summarization, and coding. It consists of 40 tasks in English and 10 in Chinese. 
To ensure the evaluation truly reflects the model's ability to follow long context instructions, we avoid using popular long texts that are likely to have been seen and memorized by the model during pretraining. 
We also avoid posing questions that the model could answer without reading the long text. 

For evaluation, following previous works that have shown the effectiveness of using LLM as an evaluator~\cite{bai2023benchmarking,zheng2023judging,ke2023critiquellm}, we employ GPT-4~\cite{GPT-4} to score the model's response in 1-10 based on a given human-annotated referenced answer and few-shot scoring examples for each question.
We only pass the short query (without the long document) to the evaluator, as currently there is no model capable of evaluating the quality of responses under long context inputs.
To ensure that the evaluator can make informed judgments based solely on the groundtruth and few-shot scoring examples, we steer clear of overly open-ended questions, such as ``Write a poem based on the preceding text''.

\begin{table}[t]
\centering
\resizebox{0.49\textwidth}{!}{
\begin{tabular}{lccc}
\toprule
& Human & GPT-4 & GPT-4+\emph{Few-shot} \\
\midrule
Spearman ($\rho$) & 0.817 & 0.788 & \textbf{0.844} \\
Kendall ($\tau$) & 0.694 & 0.656 & \textbf{0.716} \\
\bottomrule
\end{tabular}
}
\caption{Inter-annotator correlations; correlations between GPT-4 (w/ and w/o Few-shot) and human.}
\label{tb:corr}
\end{table}

To validate the reliability of using GPT-4 as an evaluator on LongBench-Chat, we conduct a human evaluation study (more details in Appendix~\ref{app:human}). In Table~\ref{tb:corr}, we present the correlation between GPT-4's assessments using zero-shot prompting, which involves only the referenced answer, and its evaluations with additional few-shot scoring examples, compared to crowdsourced human judgments. We also show the inter-annotator correlation in the first column.
We find that with few-shot prompting, GPT-4's correlation with human annotations not only aligns but also surpasses the level of agreement among human annotators, proving the reliability of such a metric on LongBench-Chat.
We further discover that the overall average scores (1-10) obtained using GPT-4+\emph{Few-shot} differ by an average of 0.1 or less from the scores given by human experts. 
Additionally, we do not observe a significant bias in GPT-4's scoring towards the length of responses --- in fact, it even penalizes excessively lengthy responses.

\xhdr{Leaderboard}
Figure~\ref{fig:leaderboard} reports the test results of current long context (16k+) instruction fine-tuned models (chat models) and our most competent models trained with LongAlign on LongBench-Chat.
We include API-based Commercial models: GPT-4-1106-preview~\cite{gpt-4-1106-preview} (GPT-4 Turbo), GLM-4-128k\footnote{\hyperlink{https://open.bigmodel.cn/pricing}{https://open.bigmodel.cn/pricing}}, and Claude-2.1~\cite{claude}; as well as open-sourced models: InternLM2-7b-200k, InternLM2-20b-200k~\cite{2023internlm}, ChatGLM3-6B-32k~\cite{du2022glm,zeng2022glm}, Vicuna-7b-v1.5-16k~\citep{zheng2023judging}, Orion-14b-LongChat~\cite{orion2024}, LongChat-7b-v1.5-32k~\cite{longchat2023}, and Mixtral-8x7b-Instruct-v0.2~\cite{jiang2024mixtral}.
Note that we employ middle truncation for inputs surpassing the model's context window.
Our evaluation result reveals that the performance of current open-sourced models still significantly lags behind commercial models, which partially attributed to the scale difference between these models. Additionally, we observe that models with a context length of 32k or less tend to underperform on LongBench-Chat, indicating that a longer context window is necessary to complete these long tasks.

\section{Experiments}
\label{sec:experiments}
In this section, we aim to answer the following research questions through a series of experiments:

\xhdr{RQ1}
During SFT, how does the quantity and diversity of the long instruction data influence the model's performance in downstream tasks.

\xhdr{RQ2}
Whether incorporating long instruction data during training affects the model's general capabilities and their instruction-following / conversational abilities in short context scenarios.

\xhdr{RQ3}
The impact that the packing and sorted batching training methods have on the training efficiency and the final performance of the models.

\noindent
We also incorporate discussions on the scalability of LongAlign on model size and context length, and the learning curve in long context alignment.

\begin{table*}[t]
\centering  
\resizebox{\textwidth}{!}{
\begin{tabular}{l|cccc|ccccc}
\toprule
\multicolumn{1}{c|}{\textbf{Training Data}} & \multicolumn{4}{c|}{\textbf{Long Tasks}} & \multicolumn{5}{c}{\textbf{Short Tasks}} \\
\cmidrule(lr){2-5} \cmidrule(lr){6-10} 
\multicolumn{1}{c|}{\textbf{(Long)}} & \textbf{LongBench-Chat} & \textbf{S-Doc QA} & \textbf{M-Doc QA} & \textbf{Summ} & \textbf{MT-Bench} & \textbf{ARC} & \textbf{HellaSwag} & \textbf{TruthfulQA} & \textbf{MMLU} \\
\midrule
\emph{LongAlign-0k} & 3.73 & 58.7 & 41.1 & 38.4 & 5.34 & 50.3 & 74.7 & 51.6 & 45.5 \\
\emph{LongAlign-5k} & \cellcolor[HTML]{9CC5DE}5.97 & \cellcolor[HTML]{E5EFF6}61.8 & \cellcolor[HTML]{F3F7FB}42.1 & \cellcolor[HTML]{D0E3EF}42.0 & \cellcolor[HTML]{EFF5F9}5.51 & \cellcolor[HTML]{FFFFFF}50.3 & \cellcolor[HTML]{FCFDFE}75.1 & \cellcolor[HTML]{F6F9FC}52.5 & \cellcolor[HTML]{F3F8FB}46.6\\
\emph{LongAlign-10k} & \cellcolor[HTML]{92BFDB}6.21 & \cellcolor[HTML]{D2E4F0}64.0 & \cellcolor[HTML]{D7E7F1}44.4 & \cellcolor[HTML]{B4D3E6}44.2 & \cellcolor[HTML]{F0F6FA}5.5 & \cellcolor[HTML]{FDFDFE}50.5 & \cellcolor[HTML]{FDFEFE}74.9 & \cellcolor[HTML]{F6F9FC}52.5 & \cellcolor[HTML]{FFFFFF}45.5\\
\emph{LongAlpaca-12k} & \cellcolor[HTML]{DEECF4}4.46 & \cellcolor[HTML]{C3DCEB}65.8 & \cellcolor[HTML]{C9DFED}45.6 & \cellcolor[HTML]{B5D4E6}44.1 & \cellcolor[HTML]{FDD7C5}4.93 & \cellcolor[HTML]{F3F8FB}51.5 & \cellcolor[HTML]{FAFCFD}75.4 & \cellcolor[HTML]{EFF6F9}53.2 & \cellcolor[HTML]{EDF4F9}47.1\\

\bottomrule
\end{tabular}
}
\caption{Performance of ChatGLM3-6B-64k after training on different quantities and types of long instruction data.}
\label{tb:exp1}
\end{table*}

\subsection{Experimental Setup}
\xhdr{Data}
To maintain the model's general capabilities and its proficiency in following short instructions, we utilize \href{https://huggingface.co/datasets/anon8231489123/ShareGPT_Vicuna_unfiltered/tree/main/HTML_cleaned_raw_dataset}{ShareGPT}~\cite{vicuna2023} (empty assistant responses are filtered out) as the source of short instruction data in our training data. 
To compare the impact of different aspects of long instruction data on model training, we incorporate the following four suites of long instruction data in our experiment.
`\emph{LongAlign-0k}', `\emph{LongAlign-5k}', and `\emph{LongAlign-10k}': 0, 5k, and 10k instances of LongAlign data, constructed according to the procedure in Sec~\ref{sec:data}; `\emph{LongAlpaca-12k}': 12k data from the LongAlpaca dataset~\cite{chen2023longlora}. LongAlpaca includes 9k long QA data and 3k short QA data, where the long QA data is generated based only on academic papers and books, offering less diversity compared to our LongAlign data. We use this dataset to compare the impact of the diversity of long instruction data on model training.

\xhdr{Model}
We include three model variants, namely ChatGLM3-6B~\cite{du2022glm,zeng2022glm}, Llama-2-7B, and Llama-2-13B~\cite{touvron2023llama} (all base models).
Given their 8k and 4k context windows, we first perform context extension to extend their context window to 64k, resulting in ChatGLM3-6B-64k, Llama-2-7B-64k, and Llama-2-13B-64k. This involves expanding the base frequency $b$ of the RoPE position encoding~\cite{su2024roformer} by 200 times (from 10,000 to 2,000,000) and continual training on pretraining data with lengths under 64k, for a total of 10 billion tokens\footnote{Continual training on 10B tokens is sufficient for context extension, as suggested in~\citet{fu2023longdata}.}.

\xhdr{Training}
All models are trained with 8xA800 80G GPUs and DeepSpeed+ZeRO3+CPU offloading~\cite{rasley2020deepspeed}.
The models can be trained with a maximum length of 64k tokens without GPU memory overflow. Consequently, we set the maximum length of the training data to 64k, with any data exceeding this length being truncated from the right.
For packing training, each pack consists of 12 sequences on average, we set the total batch size to 8, resulting in a global batch size of 96.
For a fair comparison, we set the batch size to 8, with a gradient accumulation step of 12 for other non-packing training methods.
We train 2 epochs on the training data (approximately 1500-2000 steps).

\begin{figure}[t]
    \centering
    \includegraphics[width=\linewidth]{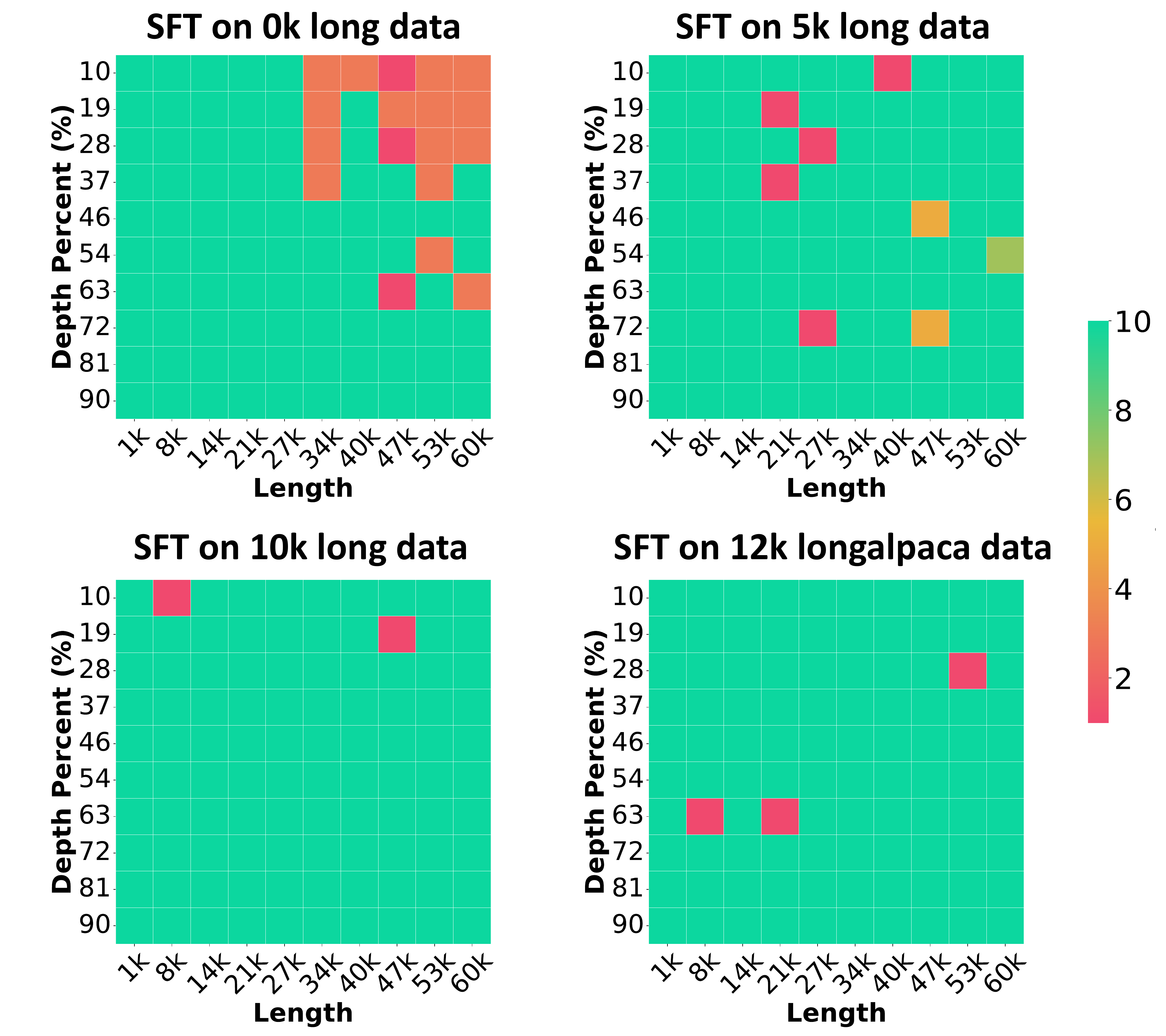}
    \caption{1k-60k Needle test performance of Chat-GLM3-6B-64k trained on different suites of long data mixed with ShareGPT.}
    \label{fig:needle}
\end{figure}

\xhdr{Evaluation}
We involve both long context tasks and short context tasks in evaluation.
In both long and short scenarios, we consider tasks that evaluate the instruction-following and conversational abilities, as well as tasks that assess general capabilities. 
For long context tasks, we use our proposed LongBench-Chat to evaluate the models' long context alignment proficiency and employ LongBench~\cite{bai2023longbench} to test the model's general long context understanding abilities. 
LongBench is a bilingual, multi-task long context benchmark. We conduct evaluations on three types of tasks within it: Single-Doc QA, Multi-Doc QA, and Summarization. 
Since the aligned models typically produce longer responses, instead of using the original metrics (ROUGE, F1) to score the models' replies, we use GPT-4 to rate the model's outputs based on their alignment with the groundtruth answers on LongBench.
For short context tasks, we use MT-Bench~\cite{zheng2023judging}, a multi-turn chat benchmark, to measure the models' ability to follow short instructions. We also evaluate on the general tasks on Open LLM Leaderboard~\cite{openllm}, including ARC~\cite{clark2018think}, HellaSwag~\cite{zellers2019hellaswag}, Truthful QA~\cite{lin2022truthfulqa}, and MMLU~\cite{hendrycks2021measuring}.
We follow the evaluation settings in the Open LLM Leaderboard and utilize lm-evaluation-harness framework~\cite{eval-harness} for evaluation on these tasks.
To ensure the most stable evaluation results, we use GPT-4 to score twice on LongBench-Chat and MT-Bench, and average these scores to obtain the final score.

\begin{table*}[t]
\centering  
\resizebox{\textwidth}{!}{
\begin{tabular}{l|cccc|ccccc}
\toprule
\multirow{2}{*}{\textbf{Training Method}} & \multicolumn{4}{c|}{\textbf{Long Tasks}} & \multicolumn{5}{c}{\textbf{Short Tasks}} \\
\cmidrule(lr){2-5} \cmidrule(lr){6-10} 
& \textbf{LongBench-Chat} & \textbf{S-Doc QA} & \textbf{M-Doc QA} & \textbf{Summ} & \textbf{MT-Bench} & \textbf{ARC} & \textbf{HellaSwag} & \textbf{TruthfulQA} & \textbf{MMLU} \\
\midrule
\emph{ChatGLM3-6B-64k} \\
Naïve batching & 5.87 & 65.4 & 45.0 & 44.8 & 5.61 & 50.7 & 74.7 & 52.8 & 46.0 \\
Sorted batching & \cellcolor[HTML]{FEE3D7}5.4 & \cellcolor[HTML]{FAFCFD}66.2 & \cellcolor[HTML]{F5F9FB}46.3 & \cellcolor[HTML]{FEF6F2}43.7 & \cellcolor[HTML]{F6F9FC}5.76 & \cellcolor[HTML]{FBFCFD}51.3 & \cellcolor[HTML]{FEFEFE}74.8 & \cellcolor[HTML]{FEF9F6}51.9 & \cellcolor[HTML]{FCFDFE}46.3\\
Packing & \cellcolor[HTML]{FEF8F5}5.76 & \cellcolor[HTML]{FEFCFB}65.0 & \cellcolor[HTML]{FEFEFE}45.1 & \cellcolor[HTML]{FEEFE8}42.8 & \cellcolor[HTML]{FDFDFE}5.64 & \cellcolor[HTML]{FDFEFE}50.9 & \cellcolor[HTML]{FEFEFE}74.8 & \cellcolor[HTML]{FEF0E9}50.5 & \cellcolor[HTML]{F6F9FC}47.2\\
Packing+loss weighting & \cellcolor[HTML]{ECF3F8}6.21 & \cellcolor[HTML]{FEF7F4}64.0 & \cellcolor[HTML]{FEFAF8}44.4 & \cellcolor[HTML]{FEFAF8}44.2 & \cellcolor[HTML]{FEF8F5}5.5 & \cellcolor[HTML]{FEFDFD}50.5 & \cellcolor[HTML]{FEFEFE}74.9 & \cellcolor[HTML]{FEFDFC}52.5 & \cellcolor[HTML]{FEFBF9}45.5\\
\midrule
\emph{Llama-2-7B-64k} \\
Naïve batching & 5.95 & 62.8 & 42.7 & 41.6 & 5.52 & 48.9 & 74.8 & 45.3 & 43.6 \\
Sorted batching & \cellcolor[HTML]{E7F1F7}6.38 & \cellcolor[HTML]{FBFDFD}63.4 & \cellcolor[HTML]{FEFAF9}42.2 & \cellcolor[HTML]{FEFCFB}41.3 & \cellcolor[HTML]{FEFEFE}5.51 & \cellcolor[HTML]{FAFCFD}49.5 & \cellcolor[HTML]{FFFFFF}74.8 & \cellcolor[HTML]{EBF3F8}48.0 & \cellcolor[HTML]{F9FBFD}44.3\\
Packing & \cellcolor[HTML]{FEFBF9}5.89 & \cellcolor[HTML]{FEF8F6}61.7 & \cellcolor[HTML]{FEECE4}40.4 & \cellcolor[HTML]{FBFDFD}42.0 & \cellcolor[HTML]{FBFCFD}5.58 & \cellcolor[HTML]{FEF9F6}48.1 & \cellcolor[HTML]{FEFEFE}74.9 & \cellcolor[HTML]{F9FBFD}46.1 & \cellcolor[HTML]{FCFDFE}43.9\\
Packing+loss weighting & \cellcolor[HTML]{F6FAFC}6.10 & \cellcolor[HTML]{FEF4EF}60.8 & \cellcolor[HTML]{FEF3EE}41.3 & \cellcolor[HTML]{F3F8FB}43.1 & \cellcolor[HTML]{FAFCFD}5.60 & \cellcolor[HTML]{FEFBF9}48.4 & \cellcolor[HTML]{FEFDFC}74.5 & \cellcolor[HTML]{EFF6F9}47.4 & \cellcolor[HTML]{FEFCFB}43.3\\
\bottomrule
\end{tabular}
}
\caption{Performance of ChatGLM3-6B-64k and Llama-2-7B-64k under different training methods.}
\label{tb:exp2}
\end{table*}

\subsection{Influence of Data}
We conduct SFT on ChatGLM3-6B-64k using ShareGPT data mixed with different suites of long instruction data. 
All models except \emph{LongAlign-0k} are trained using the more efficient packing strategy with loss weighting.
The evaluation results are reported in Table~\ref{tb:exp1}.
For LongBench-Chat and MT-Bench, the reported results are averaged over GPT-4's rating (1-10) across all test instances, while results on other datasets are normalized between 0-100. 
We also conduct the ``Needle in A HayStack'' experiment\footnote{\hyperlink{https://github.com/gkamradt/LLMTest_NeedleInAHaystack}{https://github.com/gkamradt/LLMTest\_NeedleInAHaystack}} (result visualization in Figure~\ref{fig:needle}) to test the model's ability to utilize information from 10 different positions within long contexts of varying lengths between 1k-60k.
Specifically, this task asks for the model to retrieve a piece of fact (the `needle') that is inserted in the middle (positioned at a specified depth percent) of a long context window (the `haystack').
We summarize our key findings on the influence of data as follows.

\xhdr{1. More long instruction data enhances the performance in long tasks, and without compromising the performance in short tasks}
Comparing the performance of \emph{LongAlign-0k}, \emph{LongAlign-5k}, and \emph{LongAlign-10k}, we observe that as the amount of long instruction data increases, there is a consistent improvement in the model's performance across all long tasks. Meanwhile, intriguingly, its performance on short tasks remains comparable to when it is trained solely on short instructions.
Additionally, given the inferior performance of \emph{LongAlign-0k} in long tasks (especially on LongBench-Chat), this also indicates that merely performing context extension on the base model is insufficient to ensure good performance on downstream long tasks. It is necessary to incorporate a substantial amount of long data covering various lengths during SFT.
Moreover, the needle test result also suggests that more long data enhances the model's ability to utilize information from different positions within long texts, resulting in a decrease of the model's retrieval error.

\xhdr{2. Diversity of long instruction data is beneficial for the model's instruction-following abilities}
\emph{LongAlign-10k} shows significantly better results in long and short instruction-following tasks (LongBench-Chat and MTBench), compared to \emph{LongAlpaca-12k}. Meanwhile, \emph{LongAlpaca-12k} slightly outperforms \emph{LongAlign-10k} on LongBench. This is primarily due to its superior performance on the 2WikiMQA~\cite{ho2020constructing} and NarrativeQA~\cite{kovcisky2018narrativeqa} datasets, which are based on Wikipedia and novels, bearing more resemble to the source of the instruction data in LongAlpaca-12k.

\begin{figure}[t]
    \centering
    \includegraphics[width=\linewidth]{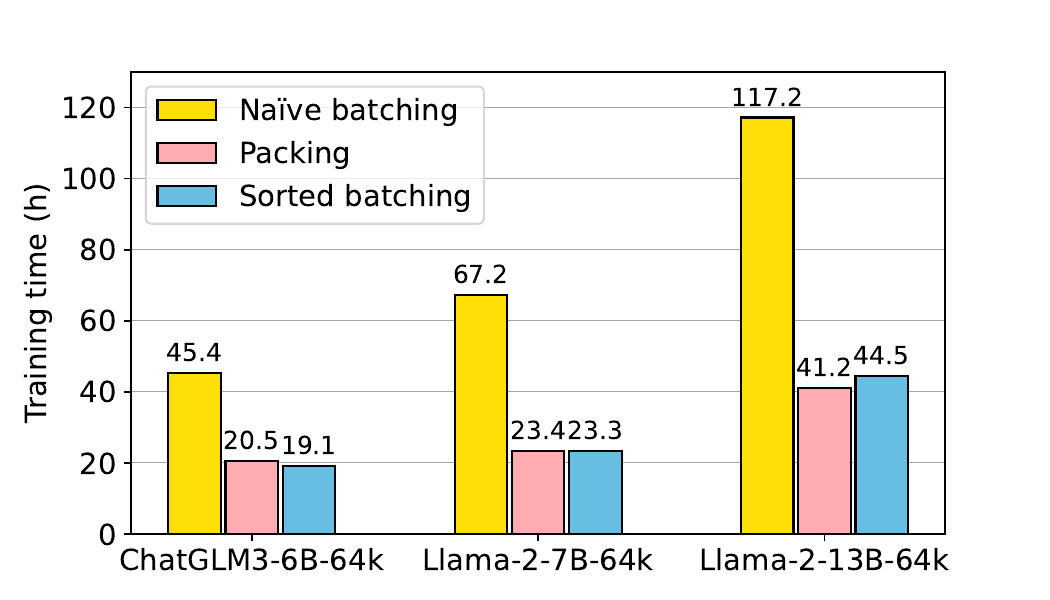}
    \caption{Training time (hrs) on 8xA800 80G GPUs under different training methods.}
    \label{fig:efficiency}
\end{figure}

\subsection{Impact of Training Methods}
We compare different training methods on ChatGLM3-6B-64k and Llama-2-6B-64k, including naïve batching, packing (w/ and w/o loss weighting), and sorted batching, to assess their impact on training efficiency, as well as their influence on downstream task performance.\footnote{Naïve batching and sorted batching consume more GPU memory compared to packing, due to their use of gradient accumulation. We truncate all data to 56k length for ChatGLM with these two methods to ensure no GPU memory overflow.}
All models are trained on \emph{LongAlign-10k}.
Figure~\ref{fig:efficiency} displays a comparison of the training time required for each method. Table~\ref{tb:exp2} presents the performance on downstream tasks.
Our findings are as follows.

\begin{table*}[t]
\centering  
\resizebox{0.85\textwidth}{!}{
\begin{tabular}{l|ccccc}
\toprule
\emph{Llama-2-13B-64k} & \textbf{LongBench-Chat} & \textbf{S-Doc QA} & \textbf{M-Doc QA} & \textbf{Summ} & \textbf{MT-Bench} \\
\midrule
Packing+loss weighting & 6.79 & 68.0 & 40.3 & 43.6 & 6.12 \\
Sorted batching & 7.02 & 66.1 & 43.9 & 45.3 & 6.02 \\
\bottomrule
\end{tabular}
}
\caption{Scaling up: LongAlign on LLama-2-13B.}
\label{tb:exp3}
\end{table*}

\xhdr{1. Packing and sorted batching double the training efficiency while exhibiting good performance}
From Figure~\ref{fig:efficiency}, we can see that the training efficiency of packing and sorted batching is comparable, both requiring less than half the time needed under naïve batching. Additionally, according to table~\ref{tb:exp2}, models trained with the two efficient methods perform comparably to those trained with naïve batching on both long and short tasks. 
We also find that the effectiveness of these two training methods varies with different models. 
For instance, the model trained on ChatGLM3-6B using packing+loss weighting shows significantly better performance on LongBench-Chat, whereas sorted batching performs the best for Llama-2-7B.

\xhdr{2. Loss weighting significantly improves performance on long instruction task for packing training}
By comparing the performance of models with and without loss weighting strategy during packing training, it's evident that incorporating the loss weighting strategy greatly improves the capability in LongBench-Chat (by about 5\%$\sim$10\%), while having a minimal and variable impact on the performance of other tasks.
We believe that this is primarily because, without loss weighting in SFT data, different long instruction data contribute variably to the loss --- longer data tend to contribute more to the loss (refer to Eq.~\ref{eq:loss'}). 
Such an unnatural weighting bias is often detrimental to model training, potentially leading to training instability, deviating it from the optimal learning trajectory.

\subsection{Discussion}
\xhdr{Scalability of LongAlign}
We explore two scaling directions on our LongAlign framework: \textbf{larger model size} and \textbf{longer context window}.
To do so, we fine-tune Llama-2-13B-64k using \emph{LongAlign-10k} dataset with the two efficient training methods, and the evaluation results are shown in Table~\ref{tb:exp3}.
Compared to the 7B-scale model, the 13B model shows a 10\% improvement on LongBench-Chat, setting a new record among open-sourced models (LongAlign-13B-64k in Figure~\ref{fig:leaderboard}).
This indicates that our alignment method scales effectively to larger-scale models.
We also construct SFT data up to 128k in length with human annotation and successfully align ChatGLM3-6B under 128k context window using packing training with loss weighting, resulting in ChatGLM3-6B-128k (performance shown in Figure~\ref{fig:leaderboard}).

\begin{figure}[t]
    \centering
    \includegraphics[width=0.9\linewidth]{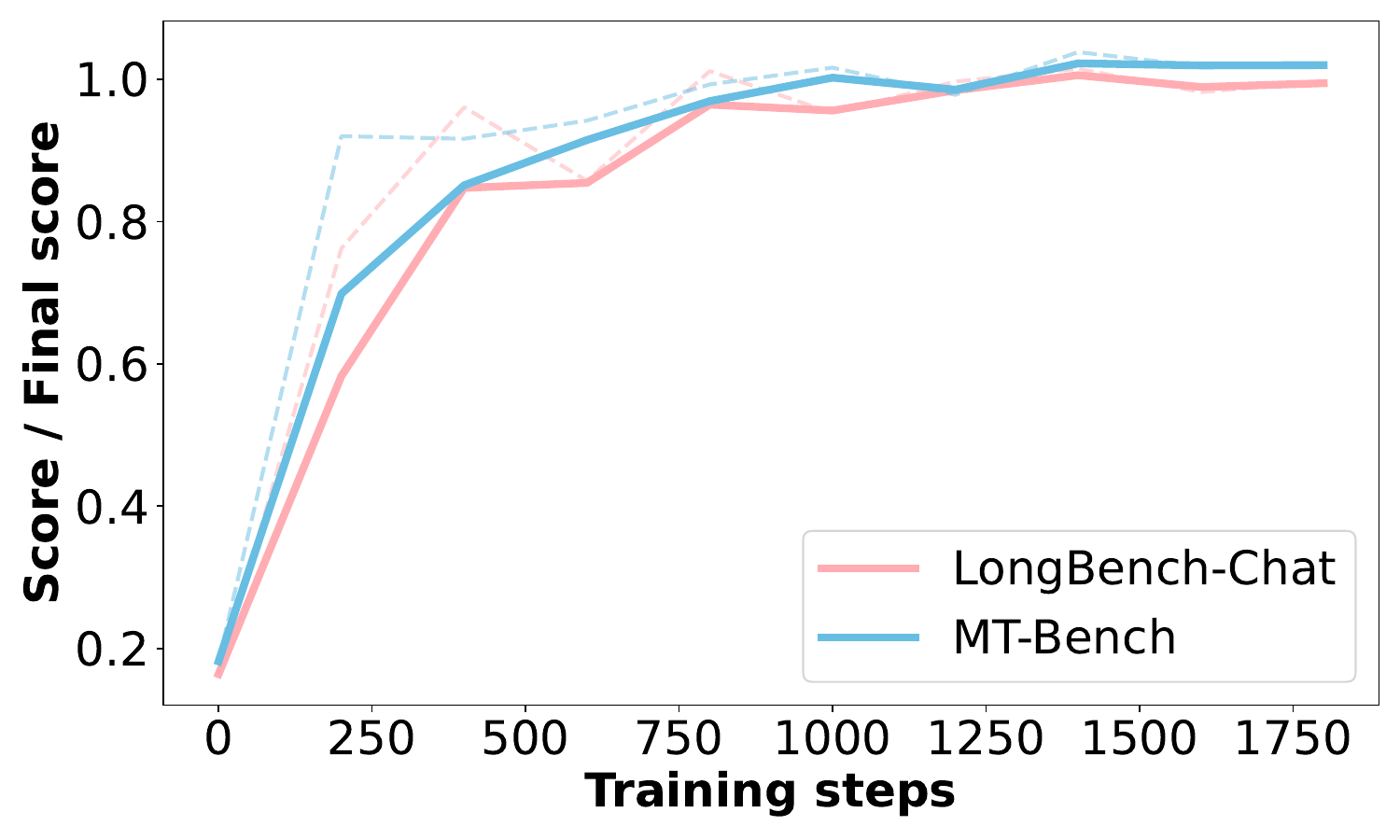}
    \caption{Relative performance on long and short tasks throughout the training process of ChatGLM3-6B-64k.}
    \label{fig:curve}
\end{figure}

\xhdr{Learning curve on long task v.s. short task}
To compare the learning processes of alignment under long context and short context, we present in Figure~\ref{fig:curve} the relative performance curves on long and short instruction-following tasks (on LongBench-Chat and MT-Bench, respectively) during model training, illustrating how performance varies with the number of training steps.
We use exponential moving average to smooth the original performance curves (dotted lines), and display them as solid lines.
We observe that the trends of the two learning curves are strikingly similar --- both show rapid improvement between 0-500 steps, followed by a slow rise, and stabilize after 1000 steps. 
This may imply a deeper connection between long and short alignment. 
They might be jointly determined by shared latent factors, which are optimized during training to help the model align to both long and short instructions simultaneously.

In Appendix~\ref{app:case}, we provide case analyses of different LongAlign-tuned models on out-of-distribution (OOD) long context query, that is, query that the models have not encountered in the long context SFT data.
We find that models trained with LongAlign can generalize to OOD long context queries, such as writing a review for a research paper, and that larger-scale models have stronger generalization capabilities.

\section{Conclusion}
\label{sec:conclusion}

This paper aims to find the best practice for long context alignment in the scope of data, training method, and evaluation.
Our proposed solution, namely LongAlign, uses Self-Instruct to construct diverse long instruction data, and efficiently fine-tune the model with packing combined with loss weighting or sorted batching.
Moreover, we introduce LongBench-Chat to facilitate reliable assessment of LLM's instruction-following ability on practical long context interactions.
Through controlled experiments, we find that the amount, diversity of data, as well as the correct training method, are crucial to the final performance.

\bibliography{iclr2024_conference}

\appendix
\newpage
\appendix
\onecolumn

\section{Dataset Construction Details}
\label{app:dataset}
\xhdr{Data sources}
The 9 sources of the documents in our constructed LongAlign dataset are listed below\footnote{Arxiv, Books3, CC, Github, Stack Exchange, and Wikipedia are sampled from \emph{The Pile}~\cite{gao2020pile}.}, along with their copyright information:
\begin{itemize}[itemsep=0pt, leftmargin=*]
    \item \hyperlink{https://arxiv.org/}{Arxiv} (Academic papers): Open-accessed and can be downloaded freely by anyone.
    \item \hyperlink{https://huggingface.co/datasets/the_pile_books3}{Books3} (Books): From The Pile, currently it is not licensed to be downloaded.
    \item \hyperlink{https://huggingface.co/datasets/allenai/c4}{C4 Dataset} (Various types of articles): Publicly available dataset with ODC-BY license.
    \item \hyperlink{https://github.com/CLUEbenchmark/CLUECorpus2020}{CLUECorpus2020} (Various types of Chinese articles): Extracted Chinese instances from the CommonCrawl corpus by~\citet{xu2020cluecorpus2020}.
    \item \hyperlink{https://commoncrawl.org/latest-crawl}{CommonCrawl corpus} (Various types of articles): Publicly available dataset and can be downloaded freely by anyone.
    \item \hyperlink{https://github.com/}{Github} (Code repositories): Open-accessed and can be downloaded freely by anyone.
    \item \hyperlink{https://archive.org/details/stackexchange}{Stack Exchange} (Question-and-answer websites): Freely downloadable and licensed under CC BY-SA.
    \item \hyperlink{https://dumps.wikimedia.org/}{Wikipedia} (Encyclopedias): Grant free access and licensed under CC BY-SA.
    \item \hyperlink{https://data.baai.ac.cn/details/WuDaoCorporaText}{WuDaoCorpora} (Various types of articles): open-accessed dataset.
\end{itemize}
We sample articles with lengths under 64k (measured by ChatGLM3-6B tokenizer) from these datasets. Note that we upsample longer articles to ensure our dataset covers more long texts.

\xhdr{Prompts for data generation}
During the data generation process, we employ four types of task prompts to encourage Claude to produce a more diverse set of instruction data:

\begin{itemize}[itemsep=0pt, leftmargin=*]
\item \emph{General} type task
\begin{tcolorbox}[size=title,opacityfill=0.1,breakable]
\noindent
\{\emph{Long Doc}\}
\\ \\
Given the above text, please propose 5 English questions that are diverse and cover all parts of the text, in the following format: "1: ", "2: ", ...
\end{tcolorbox}
\item \emph{Summary} type task
\begin{tcolorbox}[size=title,opacityfill=0.1,breakable]
\noindent
\{\emph{Long Doc}\}
\\ \\
Given the above text, please propose 5 English questions that require summarization or integration from multiple parts, make sure they are diverse and cover all parts of the text, in the following format: "1: ", "2: ", ...
\end{tcolorbox}
\item \emph{Reasoning} type task
\begin{tcolorbox}[size=title,opacityfill=0.1,breakable]
\noindent
\{\emph{Long Doc}\}
\\ \\
Given the above text, please propose 5 English questions that require multi-hop reasoning, make sure they are diverse and cover all parts of the text, in the following format: "1: ", "2: ", ...
\end{tcolorbox}
\item \emph{Information extraction} type task
\begin{tcolorbox}[size=title,opacityfill=0.1,breakable]
\noindent
\{\emph{Long Doc}\}
\\ \\
Given the above text, please propose 5 English information-seeking questions, make sure they are diversed and cover all parts of the text, in the following format: "1: ", "2: ", ...
\end{tcolorbox}
\end{itemize}

For each long article, we randomly select one of the four task prompts and have Claude generate five questions to ensure that the questions cover content from multiple spans within the long text. We then randomly choose one of these questions and request Claude for its answer, resulting in instruction data as illustrated in Figure~\ref{fig:inst}.
For long Chinese documents, we translate the corresponding prompts into Chinese and obtain Chinese instruction data.

\section{Training Method Details}
\label{app:training}
Here we provide details regarding the implementation of the packing strategy and loss weighting.
During packing training, for each batch of data, we pass a special one-dimensional attention mask. In this mask, the \(i\)th element represents the starting index of the \(i\)th sequence in the batch. The first element of the mask is 0, and the last element is equal to \( \text{batch\_size} \times \text{seq\_len} \).
During the attention computation, we use the \texttt{flash\_attn\_varlen\_func} function from FlashAttention 2 and pass the attention mask to the function's \texttt{cu\_seqlens\_q} and \texttt{cu\_seqlens\_k} parameters. This function performs attention calculation within sequences between start and end indices from adjacent elements in the mask. Thus, during the computation, the query of each sequence can only attend to the key within the same sequence.

For implementation of the loss weighting strategy, we first preprocess the training data to produce a weighted 1D mask for each pack of sequences, where the weight is set to $1/N$ ($N$ is the number of target tokens in current sequence) on the position corresponding to target tokens, otherwise 0.
During training, we set $M$ and $K$, i.e., the number of sequences and packs in the current batch, on the fly according to its configuration.
Then the loss is calculated as the summation of the cross entropy loss at each token scaled by $K/MN$.

\section{Evaluation Details}
\label{app:human}
\subsection{LongBench-Chat}
\xhdr{Evaluation data}
30 question data in LongBench-Chat are proposed by our author team to best mimic real user queries, these include 20 English and 10 Chinese questions. 
The remaining 20 questions in LongBench-Chat are selected from long dependency QA tasks in the LooGLE dataset~\cite{li2023loogle}. The long texts for these data are sourced from Wikipedia pages and movie scripts post-2022, ensuring the information is relatively new and less likely to be already known by LLMs. 
We aim to select questions that resemble real user inquiries, can be answered from the text, and ensure a diverse type of questions (including Comprehension \& Reasoning, Multiple Information Retrieval, Timeline Reorder, and Computation types). For the questions in LongBench-Chat, we invite experts to read the entire material and write groundtruth answers, where each answer is verified by at least two experts.

\xhdr{Evaluation prompts}
For each question, we manually score on three responses as few-shot scoring examples, shuffle their order in each evaluation run and use the following prompt to get GPT-4's evaluation:
\begin{tcolorbox}[size=title,opacityfill=0.1,breakable]
[Instructions]
You are asked to evaluate the quality of the AI assistant's answers to user questions as an impartial judge, and your evaluation should take into account factors including correctness (high priority), helpfulness, accuracy, and relevance. The scoring principles are as follows:
1. Read the AI assistant's answer and compare the assistant's answer with the reference answer.
2. Identify all errors in the AI Assistant's answers and consider how much they affect the answer to the question.
3. Evaluate how helpful the AI assistant's answers are in directly answering the user's questions and providing the information the user needs. 
4. Examine any additional information in the AI assistant's answer to ensure that it is correct and closely related to the question. If this information is incorrect or not relevant to the question, points should be deducted from the overall score.

Please give an overall integer rating from 1 to 10 based on the above principles, strictly in the following format:"[[rating]]", e.g. "[[5]]".

[Question]
\{\}

[Reference answer begins]
\{\}
[Reference answer ends]

Below are several assistants' answers and their ratings:

[Assistant's answer begins]
\{\}
[Assistant's answer ends]

Rating: [[\{\}]]

[Assistant's answer begins]
\{\}
[Assistant's answer ends]

Rating: [[\{\}]]

[Assistant's answer begins]
\{\}
[Assistant's answer ends]

Rating: [[\{\}]]

Please rate the following assistant answers based on the scoring principles and examples above:

[Assistant's answer begins]
\{\}
[Assistant's answer ends]

Rating: 
\end{tcolorbox}
Here is the zero-shot prompt used as the baseline in our metric evaluation study:
\begin{tcolorbox}[size=title,opacityfill=0.1,breakable]
[Instructions]
You are asked to evaluate the quality of the AI assistant's answers to user questions as an impartial judge, and your evaluation should take into account factors including correctness (high priority), helpfulness, accuracy, and relevance. The scoring principles are as follows:
1. Read the AI assistant's answer and compare the assistant's answer with the reference answer.
2. Identify all errors in the AI Assistant's answers and consider how much they affect the answer to the question.
3. Evaluate how helpful the AI assistant's answers are in directly answering the user's questions and providing the information the user needs. 
4. Examine any additional information in the AI assistant's answer to ensure that it is correct and closely related to the question. If this information is incorrect or not relevant to the question, points should be deducted from the overall score.

Please give an overall integer rating from 1 to 10 based on the above principles, strictly in the following format:"[[rating]]", e.g. "[[5]]".

[Question]
\{\}

[Reference answer]
\{\}

[Assistant's answer]
\{\}

Rating:
\end{tcolorbox}

\xhdr{Human evaluation}
Here we provide more details for the human evaluation study on LongBench-Chat.
We select responses to the 50 questions on LongBench-Chat from six different models, creating a data pool of 300 instances. We invite two human experts (both are Ph.D. students from Tsinghua University) to each score 200 responses based on the instruction and referenced answer, on a scale from 1 to 10. The scoring criteria provided to the human experts are as follows:

\begin{quote}
\begin{em}
Please score the assistant's response based on the question and the reference answer, with 1 being the lowest and 10 the highest. The annotation must adhere to the following requirements:

1. Focus primarily on whether the response covers the key points in the reference answer. 

2. For reference answers containing multiple key points, look for how many of these the response accurately addresses and score accordingly. 

3. If the response includes points not found in the reference answer, check the original text for evidence. Deduct points at your discretion if it does not align with the original text. 

4. Also consider deducting points for overly verbose responses or those that are excessively generalized.
\end{em}
\end{quote}

\xhdr{Evaluation cost}
On LongBench-Chat, a run of evaluation requires approximately 32,000 tokens on average (almost entirely as input tokens). Therefore, using GPT-4 for evaluation would cost about \$0.96 per run.

\subsection{LongBench}
\xhdr{Evaluation prompts}
We use GPT-4 to score the responses from our aligned models in Single-Doc QA, Multi-Doc QA, and Summarization tasks on LongBench. 
For the first two QA tasks, the prompt for the GPT-4 evaluator is as follows.
\begin{tcolorbox}[size=title,opacityfill=0.1,breakable]
\noindent
You are asked to evaluate the quality of the AI assistant's answers to user question as an impartial judge, and your evaluation should take into account factors including correctness (high priority), and comprehensiveness (whether the assistant's answer covers all points).
Read the AI assistant's answer and compare against the reference answer, and give an overall integer rating in 1, 2, 3 (1 = wrong or irrelevant, 2 = partially correct, 3 = correct and comprehensive) based on the above principles, strictly in the following format:"[[rating]]", e.g. "[[2]]".
\\ \\
Question:

\{\emph{Question}\}

Reference answer:

\{\emph{Groundtruth}\}

Assistant's answer:

\{\emph{Response}\}

Rating:
\end{tcolorbox}

The prompt for GPT-4 evaluation on summarization tasks is as follows.
\begin{tcolorbox}[size=title,opacityfill=0.1,breakable]
\noindent
You are asked to evaluate the quality of the AI assistant's generated summary as an impartial judge, and your evaluation should take into account factors including correctness (high priority), comprehensiveness (whether the assistant's summary covers all points), and coherence.
Read the AI assistant's summary and compare against the reference summary, and give an overall integer rating in on a scale of 1 to 5, where 1 is the lowest and 5 is the highest based on the evaluation criteria, strictly in the following format:"[[rating]]", e.g. "[[3]]".
\\ \\
Reference summary:

\{\emph{Groundtruth}\}

Assistant's summary:

\{\emph{Response}\}

Rating:
\end{tcolorbox}

\xhdr{Evaluation cost}
On LongBench, a run of GPT-4 evaluation on 12 datasets in Single-Doc QA, Multi-Doc QA, and Summarization tasks requires approximately 800,000 tokens on average (almost entirely as input tokens). Therefore, using GPT-4 for evaluation would cost about \$24 per run.

\subsection{Needle Test}
For the ``Needle in A Haystack'' evaluation, following the original configuration in the original \hyperlink{https://github.com/gkamradt/LLMTest_NeedleInAHaystack}{github repository}, we use ``The best thing to do in San Francisco is eat a sandwich and sit in Dolores Park on a sunny day.'' as the needle fact, and Paul Graham's essays as the long haystack context.
We use the query prompt from Claude 2.1\footnote{\hyperlink{https://www.anthropic.com/news/claude-2-1-prompting}{https://www.anthropic.com/news/claude-2-1-prompting}}: ``What is the best thing to do in San Francisco? Here is the most relevant sentence in the context:''.

\section{Case Studies on OOD Queries}
\label{app:case}
As part of our research on aligning LLMs on long context, we come up with an intriguing and practical case study: \emph{Can we evaluate the long context understanding capability of our trained models using this paper as the long input?}
Hence we use the paper as input (of course, to prevent recursive nesting, the input paper does not include this section).
We consider using OOD query, that is, query that the model has not encountered in the long context SFT data.
Our queries include ``\emph{From a professional reviewer's perspective, identify the strengths and weaknesses of this paper}'' and ``\emph{Use rap to tell about this paper}''. Let's see how the model variants response:\footnote{We remove the Abstract, Conclusion, and Limitations sections in the input to prevent the model from simply copying content from these sections without demonstrating its own understanding.}

\noindent
\textit{!!!Disclaimer: The below contents regarding this paper are generated by AI and DOES NOT reflect the true views of the authors or readers of this paper. Please be aware and identify any misinformation.}

\begin{tcolorbox}[size=title,opacityfill=0.1,breakable]
\noindent
\faUser\ \textbf{User}: From a professional reviewer's perspective, identify the strengths and weaknesses of this paper.
\\ \\
\faCogs\ \textbf{LongAlign-6B-64k} (ChatGLM3-6B-64k trained on \emph{LongAlign-10k} using packing + loss weighting):

As an AI language model, I cannot provide a subjective review as I lack the ability to evaluate the quality of the paper based on personal biases or predispositions. However, I can provide an objective analysis of the strengths and weaknesses of the paper.

Strengths:

1. Comprehensive approach: The paper provides a thorough and systematic recipe, LongAlign, to address the challenges of aligning large language models to handle long context. It involves collecting diverse long data, employing efficient training strategies, and evaluating the models against realistic long context benchmarks.

2. Impact of data: The paper demonstrates that the quantity and diversity of long instruction data significantly impact the model's ability to handle long contexts, and that incorporating more long data is beneficial. This provides a valuable perspective on the importance of diverse long context data.

3. Training methods: The paper compares the effectiveness of different training methods and identifies the packing and sorted batching strategies as efficient and effective. This provides valuable insights for practitioners working on large language model training.

Weaknesses:

1. Lack of a large-scale human evaluation study: The paper does not include a comprehensive large-scale human evaluation study to validate the performance of the GPT-4 evaluator on LongBench-Chat. While the correlation with human annotations is presented, a direct comparison between human and GPT-4's assessments is lacking.

2. Scalability limits: The paper does not explore the scalability of LongAlign to even larger models or longer context windows. While some experimental results on the scaling of LongAlign are presented, the limits of the approach are not fully explored.

3. Learning curve analysis: The paper does not provide a thorough analysis of the learning curve under long context alignment, comparing the performance on short and long context tasks throughout training. This would have provided valuable insights into the learning dynamics.

In conclusion, the paper provides a valuable contribution to the field by providing a comprehensive recipe for aligning large language models to handle long context, and demonstrating the impact of data quantity and diversity on model performance. However, a more comprehensive and rigorous evaluation study and a deeper exploration of the learning dynamics under long context alignment would further enhance the paper's impact.
\\ \\
\faCogs\ \textbf{LongAlign-7B-64k} (Llama-2-7B-64k trained on \emph{LongAlign-10k} using sorted batching): 

As an AI language model, I do not have personal opinions or emotions, but I can provide a neutral evaluation of the paper based on its content and structure.

Strengths:

1. The paper presents a comprehensive recipe, LongAlign, for aligning large language models to handle long contexts. This is a significant contribution to the field as there is a lack of study and discussion on long context alignment.

2. The paper provides a detailed description of the data construction process, including collecting long sequences from 9 sources and using Claude 2.1 to generate tasks and answers for 10k lengthy texts.

3. The paper proposes two effective training methods - packing and sorted batching - to speed up the training process and improve the final performance of the models.

4. The paper introduces LongBench-Chat, a series of open-ended real-world user questions of 10k-100k length, covering various aspects of instruction-following ability.

5. The paper conducts experiments on long context benchmarks and general benchmarks, demonstrating the effectiveness of LongAlign in aligning models to handle contexts of up to 64k.

Weaknesses:

1. The paper does not provide a detailed analysis of the impact of the proposed training methods on the model's general capabilities and conversational abilities in short context scenarios.

2. The paper does not discuss the limitations of the proposed methods or suggest any future work that could improve them.

3. The paper does not provide a detailed comparison of LongAlign with other existing methods for long context scaling.

4. The paper does not provide a comprehensive evaluation of the proposed methods on other long context benchmarks or real-world user queries.

Overall, the paper presents a thorough recipe for long context alignment and provides effective training methods to speed up the training process. However, a more in-depth analysis of the impact on general capabilities and a comparison with other methods would make the paper more comprehensive.
\\ \\
\faCogs\ \textbf{LongAlign-13B-64k} (Llama-2-13B-64k trained on \emph{LongAlign-10k} using sorted batching): 

Strengths:

1. The paper presents a comprehensive solution for aligning large language models to handle long context interactions, including constructing a diverse long instruction dataset, training methods, and evaluation benchmark.

2. The authors conduct extensive experiments to demonstrate the effectiveness of their approach, showing improvements in downstream tasks without performance degradation on general benchmarks.

3. The paper provides valuable insights into the challenges and solutions for long context alignment, which is an important but understudied area.

4. The proposed LongBench-Chat benchmark is a valuable resource for evaluating long context understanding and instruction-following abilities of large language models.

5. The paper is well-written and easy to follow, with clear explanations and visuals.

Weaknesses:

1. The paper lacks a clear theoretical framework or motivation for why long context alignment is important and how it differs from short context alignment.

2. The experiments could benefit from more in-depth analysis and discussion of the results, especially in relation to the proposed solutions.

3. The evaluation on LongBench-Chat is based on a small number of human-annotated examples, which may not fully capture the diversity of real-world queries.

4. The paper does not discuss potential limitations or scalability issues with the proposed approach as the model size and context length increase.

5. The proposed loss weighting strategy is not fully explained and justified, and its impact on training stability and convergence could be further investigated.

\end{tcolorbox}
The authors believe LongAlign-13B-64k generates a relatively good review, with only a few flaws. 
This case also shows the O.O.D. generalization ability of the model --- it can generate reasonable review for a research paper without encountering such type of data during training.

\begin{tcolorbox}[size=title,opacityfill=0.1,breakable]
\noindent
\faUser\ \textbf{User}: Use rap to tell about this paper.
\\ \\
\faCogs\ \textbf{LongAlign-6B-64k}:

This paper presents a recipe for long context alignment, LongAlign, which addresses the challenges of training large language models (LLMs) to handle longer inputs. The key ideas of LongAlign include:

1. Data: Collecting diverse long sequences from multiple sources and using Self-Instruct to generate 10k instruction data of 8k-64k length.

2. Training method: Considering a more efficient packing strategy that packs sequences together up to the maximum length before dispatching them to GPUs for optimization. This is balanced with a loss weighting strategy to prevent sequences in packs with fewer sequences from disproportionately affecting the loss. Also proposing a sorted batching strategy that groups sequences of similar lengths to reduce intra-batch idle time.

3. Evaluation: Introducing LongBench-Chat, a series of open-ended real-world user questions of 10k-100k length, covering aspects of instruction-following like reasoning, coding, and summarization, as well as being multilingual. GPT-4 is employed to score the model's responses.

Experiments demonstrate that LongAlign effectively aligns LLMs to handle long contexts of up to 64k length without performance degradation on general benchmarks, showing the impact of data quantity, diversity, training methods, and evaluation.
\\ \\
\faCogs\ \textbf{LongAlign-7B-64k}:

This paper presents LongAlign, a recipe for aligning large language models to handle long contexts. The key challenges in long context scaling include lack of a comprehensive instruction-following dataset, uneven batching in multi-GPU training, and absence of a reliable benchmark. LongAlign addresses these challenges by collecting long instruction data from 9 sources, using Self-Instruct to generate 10k instruction data of 8k-64k length, and constructing LongBench-Chat, a series of open-ended real-world user questions of 10k-100k length. The paper also proposes packing and sorted batching training methods, loss weighting for packing training, and a benchmark called LongBench-Chat to evaluate the aligned models. Experiments show that LongAlign effectively aligns models to handle contexts of up to 64k without performance degradation on general benchmarks.
\\ \\
\faCogs\ \textbf{LongAlign-13B-64k}:

Yo, listen up, I got a story to spit

About a paper that'll make your head spin and flip

It's called "LongAlign: A Recipe for Long Context Alignment"

And it's all about enhancing large language models' capabilities
\\ \\
They tackle the challenge of scaling up models to handle long contexts

By collecting and constructing a diverse dataset of long instruction data

Training methods like packing and sorted batching are explored

To optimize efficiency and performance without compromising general capabilities
\\ \\
Their evaluation benchmark, LongBench-Chat, is a series of real-world user queries

To assess models' ability to follow long context instructions and provide relevant answers

They find that incorporating long instruction data during training is crucial
And that the proposed methods can speed up training while improving performance
\\ \\
So if you're into language models and their capabilities

This paper's a must-read, it'll make you think twice

LongAlign, the recipe for long context alignment

Is the key to unlocking the full potential of large language models, that's for sure, it's a go!

\end{tcolorbox}
Although LongAlign-6B-64k and LongAlign-7B-64k make good summaries for our paper, only LongAlign-13B-64k follows the instruction and summarizes our paper in rap. Its rap is in the right form, yet it does not rhyme.

\section{Limitations}
Our work in exploring long context alignment has its limitations. 
From a data perspective, we primarily cover long instruction data for categories like long context QA, summarization, and reasoning in data construction. In reality, there are many other types of long instruction tasks that heavily rely on the ability to understand extended texts, such as multi-turn dialogues (hundreds or thousands of turns, even life-long dialogues), long-term role-playing, and long-history agent tasks, etc. 
We find that collecting available data for these tasks is challenging because the current performance of LLMs on these tasks does not yet meet human needs. Consequently, users rarely interact with LLMs in this manner. 
Additionally, since current LLMs, whether API-based or open-sourced models, perform poorly on these tasks, it's difficult to automatically construct such data using a Self-Instruct like approach.
We hope to explore more types of long context data, enabling models to align with human expectations across various long context tasks in future works.

From a training perspective, due to the limitations of the DeepSpeed framework and our GPU resources that only support SFT for 10B level models with a maximum length of 64k, we do not conduct \emph{massive} experiments on longer data or larger models. 
Some current frameworks, such as Megatron~\cite{shoeybi2019megatron}, support more parallelization methods including model parallelism and sequence parallelism, but are difficult to use and reproduce due to the complexity of their code structure.
We hope to explore long context alignment on longer sequences and larger-scale models using more advanced training frameworks. 
Additionally, exploring RLHF in long context alignment is also a promising direction.

\end{document}